# E-Pro: Euler Angle and Probabilistic Model for Face Detection and Recognition

Sandesh Ramesh, Manoj Kumar M V, Sanjay H A
Department of Information Science and Engineering
Nitte Meenakshi Institute of Technology
Bangalore- 560064, India
Sandeshr1515@gmail.com, Manojmv24@gmail.com

*Abstract*— **It is human nature to give prime importance to facial appearances. Often, to look good is to feel good. Also, facial features are unique to every individual on this planet, which means it is a source of vital information. This work proposes a framework named E-Pro for the detection and recognition of faces by taking facial images as inputs. E-Pro has its potential application in various domains, namely attendance, surveillance, crowd monitoring, biometric based authentication etc.**

**E-Pro is developed here as a mobile that aims to will aid lecturers to mark attendance in a classroom by detecting and recognizing the faces of students from a picture clicked through the app. E-Pro has been developed using Google Firebase Face Recognition APIs, which uses Euler Angles, Probabilistic Model. E-Pro has been tested on stock images and the experimental results are promising.**

## I. Introduction

Face recognition has made its presence in one form or another since the 1960s [1], and recent technological developments have led to a wide proliferation of this advancing technology. With its introduction on smartphone devices, millions of people around the world have this technology at the palm of their hands, storing and protecting valuable data. But to recognize a face, the device must first detect it. Face detection refers to the ability of the device or computer to identify the presence of a human entity within a digitized image or video.

Face detection has plenty of applications, most prominent applications are-
- Facial recognition- attempts to establish identity. The process usually works by using a computer application that captures a digital image of an individual's face and compares it to images already in the existence of a database [2].
- Prevention of retail crime- where shoplifters with a history of fraud enter a retail establishment are identified instantly or smarter advertising- where companies make advertising more targeted by making educated guesses at people's age
- Gender or forensic investigations- by identifying dead individuals through video surveillance footage data [3].

Another application that the paper focuses on is marking students' attendance. Objectives of the proposed model discussed in this paper are as follows:

1. Detection of face/s in the image by considering vital factors such as lighting, facial expression, facial posture, etc. for accurate recognition.
2. Calculation of 3-Dimensional (3D) geometrical orientation of the face using Euler Angle.
3. Calculation of percentage recognition by probabilistic decision.
4. Obtain identification results.
5. Calculate total count of faces in image.

The upcoming sections are ordered in the following pattern- Section II discusses a condensed overview of the work carried out by previous researchers in the domain of facial recognition and requirements needed for the same. Section III illustrates and describes the framework of E-Pro and Section IV tabulates the readings obtained through experimentation of the model. Section V concludes with the contribution of the paper with concise summary.

## II. Literature survey

According to the research paper titled "Face recognition/ detection by probabilistic decision-based neural network", the authors have designed a facial recognition system based on the probabilistic decision-based neural networks (PDBNN) comprising of a tri-methodology- the first step involves a face detector that detects the location of a human face in an image. With the help of an eye localizer, we analyze the position of eyes in order to evaluate and generate meaningful feature vectors. Facial features proposed included eyebrows, eyes, and nose, but did not include the mouth. The final module was the face recognizer. A hierarchical network of structures with nonlinear basis functions was adopted by the system. The whole recognition process consumed approximately one second, without the use of hardware accelerator or co-processor [4].

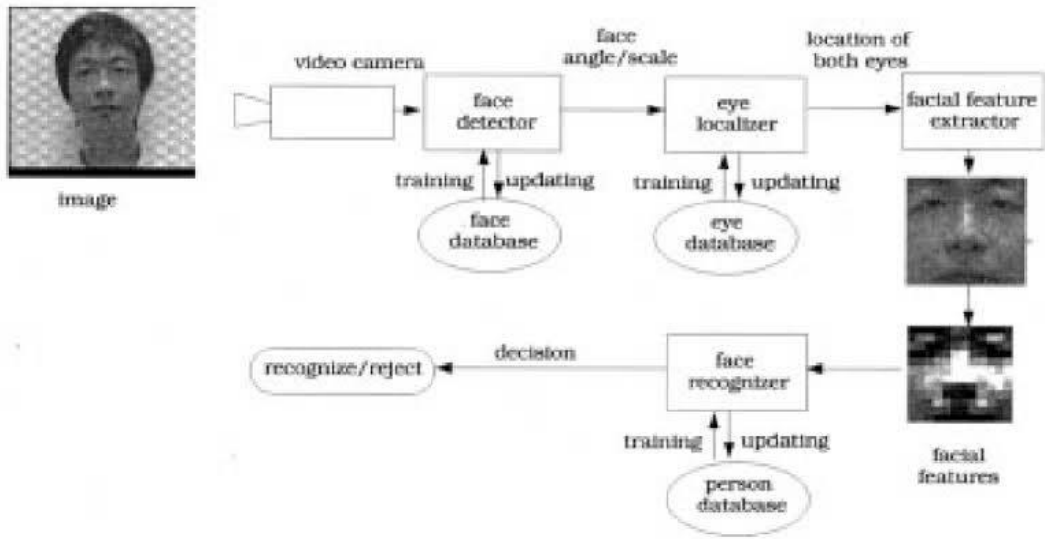

*Fig.1 Face recognition system framework as suggested by Shang- Hung Lin (2000, p.2)*

Raphaël Féraud in their paper "A Fast and Accurate Face Detector Based on Neural Networks", approached the task of detecting faces in complex backgrounds with the help of Constrained Generative Model (CGM). Generative, since the aim of the entire learning process was to evaluate the probability that the model generated from the data inputted and constrained since some counter examples were utilized to enhance the overall quality of the estimated performance by the model. Also, in order to detect the side view of faces and to reduce the occurrence of false alarms, a conditional mixture of networks was utilized [5]. In early developmental stages of facial detection algorithms, focus was mainly directed towards the frontal part of the human face. According to the research paper titled "A Neural Basis of Facial Action Recognition in Humans", the authors state that by combining various facial muscular actions, called action units, humans have the ability to produce an large number of facial expressions [6]. With the use of functional magnetic resonance imaging and certain machine learning technique, the authors were able to identify a consistent and differential coding of action units by the brain. Most of the attempts by Hjelmås and Low, have aimed to improve the ability to cope with alterations, but are still restricted to certain body parts such as head, shoulder and frontal face. There is however, a rising need for the detection of multiple faces in a clustered environment e.g. clutter-intensive background. Furthermore, the proposed method goes on to ignore the basic understanding of the face and tends to use facial patterns from stock images. This is mostly known as the training stage in the detection method [7]. They go a step further to put the system to experiment, based on a series of combination of various classifiers to help in obtaining more reliable result than compared to using a single classifier. The authors have designed multiple face classifiers that are capable of taking different representations of face patterns. The authors have used three types of classifiers, the first one being Gradient feature classifier. This feature contains the information of pixel distribution that returns certain invariability among facial features. The second classifier used is called the Texture Feature. This feature helps in extracting texture features by correlation (uses joint probability occurrence of a specified pixel), variance (measures the number of local variations in an image) and entropy (measures image disorder). The third classifier the authors have used is called the Pixel Intensity Feature. This feature helps in extracting pixel intensity of the eye, nose and mouth region for exacting a particular face pattern.

III. FRAMEWORK

Functional requirements are defined as those requirements the system must deliver. In the case of E-Pro, it was vital that we collect certain requirements that will be necessary to achieve the goals set out for this system. Both functional and non-functional requirements were collected. However, the functional requirements gathered were important, as in, without these, the entirety of this system would be a failure. These requirements were also based on the users' feedback about the overall functioning of the system. The following algorithm defines a detailed overview of the objectives we aim to achieve through E-Pro.

Step 1: Start

Step 2: Capture face images via camera.

Step 3: Detection of faces within an image must be confirmed.

Step 4: Bounding boxes must be used around each face.

Step 5: Complete attendance must be marked based on number of faces obtained.

Step 6: All detected faces must be cropped.

Step 7: Cropped images can be resized to meet mapping requirements.

Step 8: All cropped images must be stored onto a folder.

Step 9: The database must be loaded with images of faces.

Step 10: These images can be used for training the model.

Step 11: Capture image and recognize faces over and again.

Step 12: Compare image stored in the database along with the input image.

Step 13: Display the name or ID of the student over the image captured

Step 14: Stop

Other features of E-Pro include specific criteria to judge the operation of the E-Pro system. They cover ease, protection, availability of support, speed of operation, and considerations for implementation. To be more specific, the user will find it very convenient to capture photographs and inform students regarding their facial positions. The system is very secure and can be easily installed. With a response time of less than 10s, E-Pro is fast and reliable

"Not Have" – These are features that could be implemented in the future to enhance the overall working of the system.

A. *Must-Have*

With regards to this model, the "Must Have" are the requirements that have been identified by the user that must be used for obtaining the desired output. Absence of these requirements, the ultimate outcome will not be achieved.
- Bounding boxes must surround the faces of people in the image.
- Images of all faces detected must be cropped.
- Captured image sizes must be cropped to meet the image size of the database.
- The total attendance of the class must be calculated based on the number of faces detected.
- Images must be trained rigorously for recognition.
- The input and output images must be displayed side-by-side.
- Name of image outputted must be displayed above it.

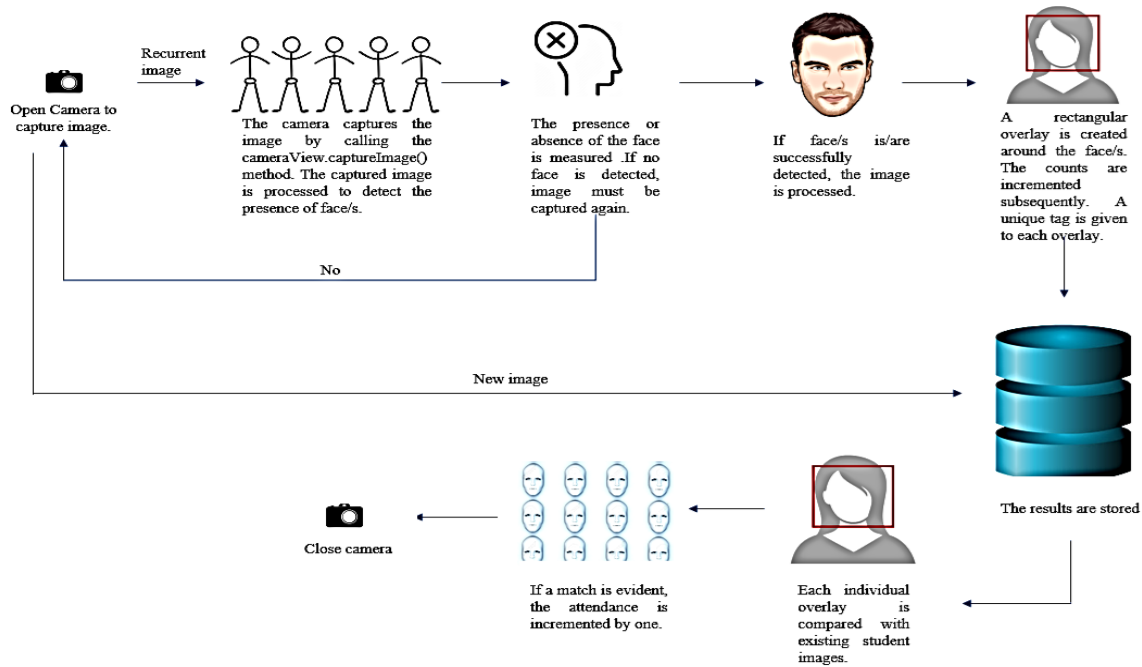

*Fig.2 Proposed E-Pro system framework*

One needs to prioritize the importance of certain requirements that are vital for the functioning of the system. To make this process easy to understand we classify the requirements under Must Have, Should Have, Could Have, and Will Not Have. "Must Have" – These are requirements without which the system ceases to exist. The "Should Have" – These are requirements with a slightly higher priority and must be implemented if possible. "Could Have" - These are features that are desirable in the system but not a necessary requirement.

B. *Should Have*

These features will be implemented if possible. These features form a priority to the system. However, even without these features, the system will continue to perform its functions.
- Ensure the names of both output and input search images are displayed.
- Calculate the recognition percentage of an image captured by the system to that of its database counterpart.

- Calculate the rate at which the system effectively recognizes faces.

## C. Could Have:

These are features, if added, could make the application much more interactive and fun to use. However, without these, the app doesn't stop functioning and continues to perform its duty.
- An enhanced and easy to use Graphical User Interface (GUI).
- High definition Camera to capture quality images.

## D. Will Not Have:

Under this header we have features that have not been included in the current system, as we don't see much use of them. Since the system is college/university specific, we find it easy to maintain the data records of students and teachers. However, should this system be used in a much larger commercial scale, the database and servers must be revamped to meet the demands of every college/university simultaneously.

## IV. METHODOLOGY

The below table contains experimental readings of people, individually and in a group of greater than two. The following abbreviations are used:

Zzwy- getBoundingBox ()- Returns the axis-aligned bounding rectangle of the detected face.

Zzxq- getTrackingId ()- Returns the tracking ID if the tracking is enabled.

Zzxu- getHeadEulerAngleY ()- Returns the rotation of the face about the vertical axis of the image.

Zzxv- getHeadEulerAngleZ ()- Returns the rotation of the face about the axis pointing out of the image.

Zzxt- getIsSmilingProbability ()- Returns a value between 0.0 and 1.0 giving a probability that the face is smiling.

Zzxs- getIsLeftEyeOpenProbability ()- Returns a value between 0.0 and 1.0 giving a probability that the face's left eye is open.

Zzxr- getIsrightEyeOpenProbability ()- Returns a value between 0.0 and 1.0 giving a probability that the face's right eye is open.

Vital factors such as lighting, facial expression, facial posture, etc. need to be considered for accurate recognition [9].

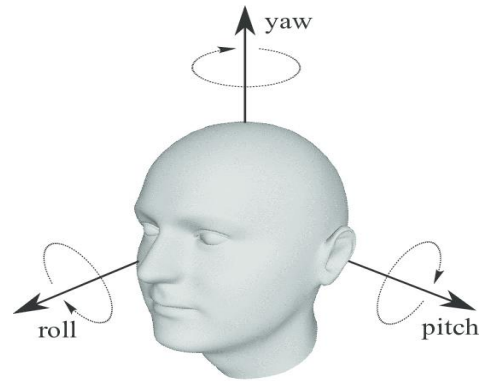

*Fig.3 Head pose angles and axes as illustrated by Jerry (2018, p22)*

## V. RESULTS AND DISCUSSIONS

The attributes obtained by detecting faces in images are tabulated below. For the purpose of experimentation, we do not take into consideration the Smiling Probability (zzxt), LeftEyeOpenProbability (zzxs), RightEyeOpenProbability (zzxr) and TrakingId (zzxq) of the images and are assigned the values -1.0, -1.0, -1.0 and 0 respectively. The rectangular bounding box around faces is represented by Rect ((left, top), (width, height)). This method indicates the size of boxes that need to be created during the process of facial identification. Groups of people are sub-divided into individual entities and their values are recorded. Observation 1 (Fig.1) takes into account, a person and a miscellaneous object, in this case a dog. The E-Pro system successfully generates a box around the person but not the object. Euler head angles (zzxu and zzxv) are those that describe the orientation of a rigid head with respect to a fixed coordinate system. Negative zzxu and zzxv values indicate that the heads are pointed in the negative direction of the axes. The system also successfully identifies a large group of people as well. Observations 5 (fig.8) and 6 (fig. 9) demonstrate this.

TABLE I. QUANTITATIVE ATTRIBUTES OF FACIAL RECOGNITION

| Observation | Number of people under consideration | Individual Classification | zzwy | zzxq | zzxu | zzxv | zzxt | zzxs | zzxr |
|---|---|---|---|---|---|---|---|---|---|
| 1 | 1 | a. P1 | Rect(270,488–469, 687) | 0 | -5.2812 | -2.9454 | -1.0 | -1.0 | -1.0 |
| | | b. Misc. | | | | | | | |

| | | | | | | | | | | |
|---|---|---|---|---|---|---|---|---|---|---|
| **2** | **2** | a. | P1 | Rect(134,494-234,593) | 0 | -3.5252 | -3.9695 | -1.0 | -1.0 | -1.0 |
| | | b. | P2 | Rect(117,872-202,9580 | 0 | -3.1259 | -0.4368 | -1.0 | -1.0 | -1.0 |
| **3** | **3** | a. | P1 | Rect(43,438– 296,697) | 0 | 36.8327 | -21.325 | -1.0 | -1.0 | -1.0 |
| | | b. | P2 | Rect(283,443– 495,655) | 0 | 1.4248 | -2.4146 | -1.0 | -1.0 | -1.0 |
| | | c. | P3 | Rect(459,491– 715,757) | 0 | -20.905 | 9.9750 | -1.0 | -1.0 | -1.0 |
| **4** | **6** | a. | P1 | Rect(86,528– 214,655) | 0 | -1.7599 | -17.752 | -1.0 | -1.0 | -1.0 |
| | | b. | P2 | Rect(480,505– 610,636) | 0 | -4.8044 | -7.6158 | -1.0 | -1.0 | -1.0 |
| | | c. | P3 | Rect(528,662– 675,813) | 0 | -5.0763 | -16.526 | -1.0 | -1.0 | -1.0 |
| | | d. | P4 | Rect(305,691– 459,844) | 0 | -11.528 | -7.7964 | -1.0 | -1.0 | -1.0 |
| | | e. | P5 | Rect(303,691– 420,608) | 0 | -1.4715 | -12.268 | -1.0 | -1.0 | -1.0 |
| | | f. | P6 | Rect(117,679– 278,836) | 0 | 5.7566 | -33.651 | -1.0 | -1.0 | -1.0 |
| **5** | **8** | a. | P1 | Rect(581,573- 665, 658) | 0 | -12.467 | 0.38612 | -1.0 | -1.0 | -1.0 |
| | | b. | P2 | Rect(498,561- 575, 640) | 0 | -9.00484 | 7.73211 | -1.0 | -1.0 | -1.0 |
| | | c. | P3 | Rect(73, 595- 151, 672) | 0 | 5.829581 | -18.794 | -1.0 | -1.0 | -1.0 |
| | | d. | P4 | Rect(279,594- 343, 659) | 0 | -1.99653 | -0.5076 | -1.0 | -1.0 | -1.0 |
| | | e. | P5 | Rect(365,567- 440, 642) | 0 | -3.63065 | -3.3822 | -1.0 | -1.0 | -1.0 |
| | | f. | P6 | Rect(212,665- 288, 741) | 0 | 8.367999 | -10.364 | -1.0 | -1.0 | -1.0 |
| | | g. | P7 | Rect(163,570- 234, 642) | 0 | 1.052805 | -5.9284 | -1.0 | -1.0 | -1.0 |
| | | h. | P8 | Rect(394,665- 475, 747) | 0 | -8.54884 | -8.5854 | -1.0 | -1.0 | -1.0 |
| **6** | **10** | a. | P1 | Rect(233,683- 300, 750) | 0 | -3.98541 | 2.441709 | -1.0 | -1.0 | -1.0 |
| | | b. | P2 | Rect(416,627- 465, 676) | 0 | 0.487392 | 0.209005 | -1.0 | -1.0 | -1.0 |
| | | c. | P3 | Rect(295,604- 348, 657) | 0 | -0.80602 | -5.32229 | -1.0 | -1.0 | -1.0 |
| | | d. | P4 | Rect(182,610- 237, 665) | 0 | -7.00699 | -1.65343 | -1.0 | -1.0 | -1.0 |
| | | e. | P5 | Rect(65, 693- 135, 764) | 0 | 1.865780 | -3.20102 | -1.0 | -1.0 | -1.0 |
| | | f. | P6 | Rect(510,612- 565, 668) | 0 | -1.32773 | -11.9088 | -1.0 | -1.0 | -1.0 |
| | | g. | P7 | Rect(61, 611- 119, 670) | 0 | 6.605692 | -0.36110 | -1.0 | -1.0 | -1.0 |
| | | h. | P8 | Rect(496,678- 570, 753) | 0 | -0.28733 | -1.65976 | -1.0 | -1.0 | -1.0 |
| | | i. | P9 | Rect(605,618- 658, 672) | 0 | 6.210866 | -10.6227 | -1.0 | -1.0 | -1.0 |

| | | | j. | P10 | Rect(380,691 - 438, 749) | 0 | 3.595529 | -5.46549 | -1.0 | -1.0 | -1.0 |

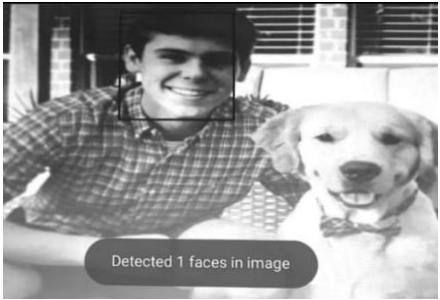
*Fig.4. Image w.r.t observation 1 in Table 1*

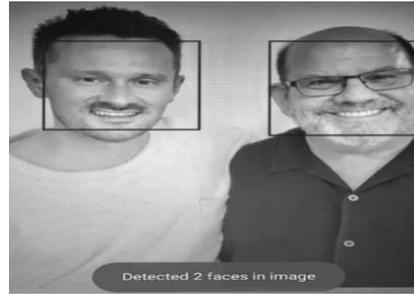
*Fig.5. Image w.r.t observation 2 in Table 1*

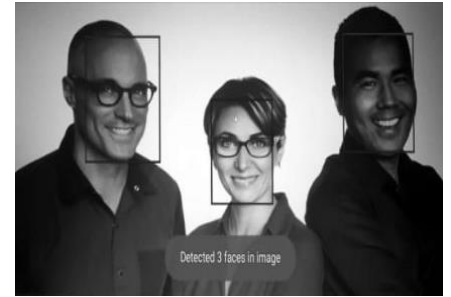
*Fig.6. Image w.r.t observation 3 in Table 1*

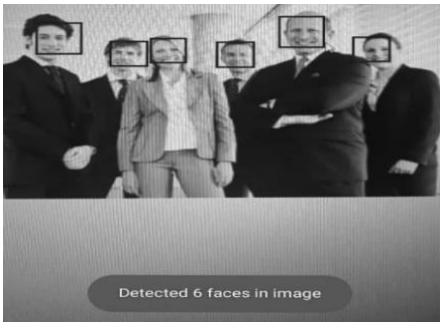
*Fig.7. Image w.r.t observation 4 in Table 1*

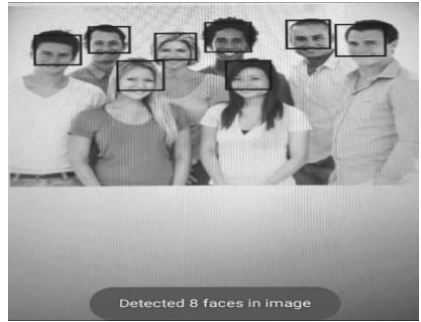
*Fig.8. Image w.r.t observation 5 in Table 1*

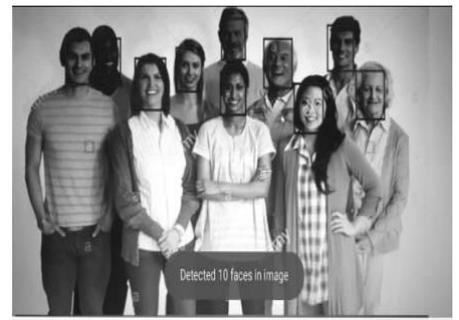
*Fig.9. Image w.r.t observation 6 in Table 1*

## VI. CONCLUSION

This paper has successfully demonstrated the use of a facial detection and recognition system named E-Pro on a dataset composing of varied number of faces. Face detection is rapidly growing in the technology sphere. While the system finds its usefulness in the attendance domain, it can also be used in the surveillance of video footages, access and security, criminal identification and payments. Any personal information can become sensitive information. With the technology's constant development, one can expect better tools for law enforcement, customized advertisements and pay-by-face authentication in the future.